\documentclass[10pt,twocolumn,letterpaper]{article}

\usepackage{cvpr}              % To produce the CAMERA-READY version
\usepackage{graphicx}
\usepackage{amsmath}
\usepackage{amssymb}
\usepackage{booktabs}
\usepackage{multirow}
\usepackage[normalem]{ulem}
\usepackage[ruled, vlined, linesnumbered]{algorithm2e}%
\usepackage{algpseudocode}
\usepackage{amsmath}
\usepackage{amsthm}
\usepackage{bbm}
\usepackage[ruled,vlined]{algorithm2e}
\usepackage{mathtools}
\graphicspath{ {latex/figures/} }

\usepackage[pagebackref,breaklinks,colorlinks]{hyperref}

\usepackage[capitalize]{cleveref}
\crefname{section}{Sec.}{Secs.}
\Crefname{section}{Section}{Sections}
\Crefname{table}{Table}{Tables}
\crefname{table}{Tab.}{Tabs.}

\newcommand*{\Comb}[2]{{}^{#1}C_{#2}}%
\def\cvprPaperID{2449} % *** Enter the CVPR Paper ID here
\def\confName{CVPR}
\def\confYear{2022}

\begin{document}
\SetKwInOut{Initialize}{Initialize}
%%%%%%%%% TITLE - PLEASE UPDATE
\title{Dynamic Kernel Selection for Improved Generalization and Memory Efficiency in Meta-learning}

% \author{Arnav Chavan \\Transmute AI Lab \and Rishabh Tiwari \and Udbhav Bamba \and Deepak Gupta
\author{Arnav Chavan$^{*}$, Rishabh Tiwari\thanks{indicates equal contribution. Arnav Chavan is the corresponding author.}$^{*}$, Udbhav Bamba and Deepak K. Gupta \\
Transmute AI Lab (Texmin Hub), Indian Institute of Technology, ISM Dhanbad \\
{\tt\small \{arnavchavan04,akchitra99,ubamba98,guptadeepak2806\}@gmail.com} \\ 
% For a paper whose authors are all at the same institution,
% omit the following lines up until the closing ``}''.
% Additional authors and addresses can be added with ``\and'',
% just like the second author.
% To save space, use either the email address or home page, not both
}
\maketitle

%%%%%%%%% ABSTRACT
\begin{abstract}
   Gradient based meta-learning methods are prone to overfit on the meta-training set, and this behaviour is more prominent with large and complex networks. Moreover, large networks restrict the application of meta-learning models on low-power edge devices. While choosing smaller networks avoid these issues to a certain extent, it affects the overall generalization leading to reduced performance. Clearly, there is an approximately optimal choice of network architecture that is best suited for every meta-learning problem, however, identifying it beforehand is not straightforward. In this paper, we present \textsc{MetaDOCK}, a task-specific dynamic kernel selection strategy for designing compressed CNN models that generalize well on unseen tasks in meta-learning. Our method is based on the hypothesis that for a given set of similar tasks, not all kernels of the network are needed by each individual task. Rather, each task uses only a fraction of the kernels, and the selection of the kernels per task can be learnt dynamically as a part of the inner update steps. \textsc{MetaDOCK} compresses the meta-model as well as the task-specific inner models, thus providing significant reduction in model size for each task, and through constraining the number of active kernels for every task, it implicitly mitigates the issue of meta-overfitting. We show that for the same inference budget, pruned versions of large CNN models obtained using our approach consistently outperform the conventional choices of CNN models. \textsc{MetaDOCK} couples well with popular meta-learning approaches such as iMAML \cite{rajeswaran2019meta}. The efficacy of our method is validated on CIFAR-fs \cite{bertinetto2018meta} and mini-ImageNet \cite{vinyals2016matching} datasets, and we have observed that our approach can provide improvements in model accuracy of up to 2\% on standard meta-learning benchmark, while reducing the model size by more than 75\%. Our code is available at \url{https://github.com/transmuteAI/MetaDOCK}.
\end{abstract}

%%%%%%%%% BODY TEXT
\section{Introduction}
% \hl{Para1 - Describe quick adaptation to unseen task in AI. Describe what is meta learning and it's advantages.}\\
% An important characteristic desired in any artificial intelligence (AI) system is \hl{that it is able} the ability to solve tasks under several different conditions, and \hl{can} adapt quickly in unseen environments. While the field of deep learning has progressed significantly with the development of several efficient algorithms, it is well known that the standard learning methods tend to break down when the statistics of the inference data differs from that of the training set \cite{}. \dpk{@Udbhav can you complete this paragraph with first paragraph of spicy opening statements?}

An important characteristic desired by any artificial intelligence (AI) system is the ability to solve tasks under several different conditions, and that it can adapt quickly in unseen environments with the help of limited data. While the field of deep learning has progressed significantly with the development of several efficient algorithms, it is well known that the standard learning methods tend to perform well when training is done with the millions of data points and break down when the statistics of the inference data differs from that of the training set. Meta-learning tries to solve this problem by \textit{learning-to-learn} the model weights over different episodes from a family of tasks with limited data. This learning process helps the models to generalize well and adapt quickly over unseen tasks with the help of few examples. Such setting has many practical advantages in vision and reinforcement learning problems like few-shot image classification \cite{vinyals2016matching, finn2017model, rajeswaran2019meta, lake2011one}, navigation \cite{wortsman2019learning}, domain adaptation \cite{guo2020broader} to name a few.

% \hl{Para2 - Describe types of meta-learning. Introduce gradient/optimization based methods. Advantages of optimization based methods vs nearest neighbours and memory based methods}\\ -- Skipping will include this in related work

% \hl{Para3 - Gradient based approaches face the problem of overfitting to a greater extent than other methods. There are two kind of overfitting involved- one is inner-task overfitting, an extensive research has been done on this problem as it commonly seen in all the deep learning approaches and several methods have been proposed to counter this eq- inner-regularization, dropout, weight decay, learning rate scheduling, etc. The second type of overfitting is inter-task overfitting or meta-overfitting, the study on this problem is limited, few methods have used pruning as method of model weight initilization \cite{}(Meta Learning in Network Pruning) and show generalization to certain extent but the fail to induce any sparsity and are very compute heavy as they require 3x the training of with respect to normal meta-learning} Even though meta-learning methods learn to adapt on the unseen tasks, they still tend to overfit on the meta-training set. This has been studied in several recent works and remedies to circumvent this issue include .....\dpk{@Udbhav: Discuss briefly here the different ways that have been proposed to improve generalization in meta-learning.} Most of these methods incline towards choosing larger CNN networks and adding implicit regularizations to avoid overfitting.

The key idea behind these kind of meta-learning approaches is to learn generalized weights which can be modified easily for a newer task. Gradient based approaches tend to face the problem of overfitting to a greater extent than other methods. The two kind of overfitting observed in them are - 1) Inner-task overfitting, which refers to task-specific overfitting of the meta-model, an extensive research has been done on this problem \cite{lee2019meta, zintgraf2019fast} as it commonly seen in all the deep learning approaches and several methods have been proposed to counter this, for example, inner-regularization, dropout, weight decay, learning rate scheduling, etc. 2) Inter-task overfitting or meta-overfitting, that is overfitting of meta-model on seen tasks and failure to generalize well on unseen tasks, the study on this problem is limited, few of the common approaches are adding implicit regularizations \cite{rajeswaran2019meta}, choosing larger CNN networks to increase learning capacity \cite{lee2019meta}, using initialization techniques to improve generalisation \cite{tian2020meta}.

% This has been studied in several recent works and remedies to circumvent this issue include .....\dpk{@Udbhav: Discuss briefly here the different ways that have been proposed to improve generalization in meta-learning.} Most of these methods incline towards choosing larger CNN networks and adding implicit regularizations to avoid overfitting.

% \dpk{Based on the above two paragraphs, we discuss the existing limitations of the methods briefly and create a research gap that is answered in the later paragraphs through our research.}

In this paper, we propose to improve the generalization of  gradient based meta-learning models as well as the associated memory efficiency through a pruning-in-learning strategy. We present \emph{\textbf{Meta}-Learning with \textbf{D}ynamic \textbf{O}ptimization of \textbf{C}NN \textbf{K}ernels} (\textsc{MetaDOCK}), a gradient-based dynamic learning scheme to identify the optimal set of kernels for each meta-learning task. \textsc{MetaDOCK} is based on the hypothesis that each task needs only a small subset of kernels from the complete set of kernels that exist in the traditional meta-learning models, and using excessive kernels could potentially cause meta-overfitting. This reduction of the number of kernels per task breaks down further into two parts: reducing the kernels in the final meta-model and learning to further optimize the choice at the task-level during the inner update steps. \textsc{MetaDOCK} uses the gradient information accumulated during the inner update steps to activate/deactivate the kernels in the meta-model.

% \dpk{Can we put an example figure and brief statistics of one of the example experiments to explain the concept? See the equivariance paper to get an idea of what I mean.}

The major contributions of \textsc{MetaDOCK} are are follows:

\begin{itemize}
    \item We demonstrate that meta-learning models can be made to generalize better on unseen tasks through efficient pruning of the excessive and undesired parts of the network at the meta-level as well as for each task.
    \vspace{-0.05in}
    \item Our \textsc{MetaDOCK} strategy dynamically identifies the right set of kernels that are to be retained for each task, and discards the rest. This helps to avoid overfitting and improves the reliability of the meta-learning \mbox{methods}.
    \vspace{-0.05in}
    \item Through pruning the meta-model as well as the task-specific models, \textsc{MetaDOCK} reduces the size of the model significantly. The resultant smaller meta-models are better suited for deployment on low-power devices and improve the inference efficiency of the model.
    \vspace{-0.05in}
    \item We demonstrate through successful integration of \textsc{MetaDOCK} with popular meta-learning approach: iMAML wherein \textsc{MetaDOCK} improves the performance on unseen tasks on benchmark datasets.
\end{itemize}

\section{Related Work}
Most standard machine learning models cannot easily adapt to unseen tasks, and meta-learning attempts to circumvent this issue \cite{DBLP:journals/corr/abs-2004-05439}. Common approaches to solve meta-learning  problems\cite{weng2018metalearning} are metric-based learning, model-based methods and optimization-based methods. In metric-based learning, the core idea is similar to nearest neighbors algorithms where an embedding function is used to encode the input signal to smaller dimensional feature vector which are further used to distinguish one class from another. These methods makes an assumption that the sample embeddings of same classes should be closer to each other and those of different classes should be further apart. Siamese Neural Network \cite{Koch2015SiameseNN}, matching networks \cite{NIPS2016_90e13578}, prototypical networks \cite{DBLP:journals/corr/SnellSZ17} are few of the prominent works in this domain. Model-Based \cite{10.5555/3045390.3045585, DBLP:journals/corr/MunkhdalaiY17, Yan2019ADA} methods design models for fast learning, this is either archived by model's internal design or with the help of another meta-model. Optimization-Based methods use a modified back-propagation to handle learning by few samples across tasks. Few of the common approaches are MAML \cite{finn2017model}, iMAML \cite{rajeswaran2019meta} and Reptile \cite{nichol2018first}. 

The method proposed in this paper is related to reducing the inter-task overfitting, also referred to as meta-overfitting in the optimization-based meta-models, with the help of task-specific kernel selection. We focus our study on the optimization-based methods as these methods are inherently designed for smaller models as compared to metric-based and model-based meta learning methods. We achieve this by developing a pruning strategy that finds different subset of kernel for each different task. Network pruning is one the the prominent ways to remove redundant weights in deep neural network. Some of the common ways to prune networks are as follows. 1) Unstructured pruning \cite{frankle2018lottery, lecun1990optimal, dong2017learning, han2015deep, han2015learning, zhu2017prune}, these methods reduce the storage requirements, however, no acceleration is currently possible through recent hardware. 2) Structured pruning \cite{tiwari2021chipnet, Li2017PruningFF, Luo2017ThiNetAF, He2017ChannelPF, Liu2017iccv}, these methods prune the entire channels or layers of the neural network maintaining the structure of the neural network. The kernel selection pruning used by our method is closely related to structured pruning approaches as it maintains the regular structure and selects the kernel to apply for each output channel in a convolutional layer. Another approach that uses pruning to improve generalization in meta-learning is proposed  in \cite{tian2020meta}. They use pruning as a method of model weight initialization and show generalization to certain extent but they fail to induce any sparsity or compression and are very compute heavy as they require 3$\times$ the training with respect to normal meta-learning without any memory/compute gain in inference . 

\section{Dynamic Kernel Selection}

\subsection{Background: Meta-learning}
In this paper, we discuss meta-learning in the context of few-shot supervised learning problems, as originally described in \cite{finn2017model}. In this setting, let $\{\mathcal{T}_i\}^M_{i=1}$ denote a set of meta-training tasks drawn from a distribution of tasks $P(\mathcal{T})$. With each task $\mathcal{T}_i$, there is a dataset $\mathcal{D}_i$ associated to it, and this dataset is further divided into two disjoint sets: $\mathcal{D}^{\text{tr}}_i$ and $\mathcal{D}^{\text{test}}_i$. The two data subsets take the form $\mathcal{D}_i^{\text{tr}} = \{(\mathbf{x}_i^k, \mathbf{y}_i^k)\}_{k=1}^K$, and similarly for $\mathcal{D}_i^{\text{test}}$, where $\mathbf{x} \in \mathcal{X}$ and $\mathbf{y} \in \mathcal{Y}$ denote input and output, respectively. The goal then is to learn models of the form $\mathcal{F}_{\phi}: \mathcal{X}\rightarrow\mathcal{Y}$, parameterized by $\phi \in \Phi$. For the task $\mathcal{T}_i$, the goal is to learn task-specific parameters $\phi_i$ using $\mathcal{D}_i^{\text{tr}}$ such that the test loss for this task, defined as $\mathcal{L}(\phi_i, \mathcal{D}_i^{\text{test}})$, is minimized.

Conventionally, meta-learning is posed as a bi-level optimization problem involving updates of the optimization parameters of the model at two different levels: meta update (outer update) and adaptation  (inner update). At the meta-update level, parameter set $\mathbf{\theta} \in \Theta$ is optimized and this together with task-specific training set are used to obtain $\phi_i$ for any given task $\mathcal{T}_i$. The goal of meta-learning is then to optimize the meta-parameters $\boldsymbol\theta$ such that for the optimal solution, its adaptation, denoted by $\mathcal{A}d(\cdot)$, for a certain task $\mathcal{T}_i$ using $D^{\text{tr}}_i$ , minimizes the test loss of the respective task, $\mathcal{L}(\boldsymbol\phi_i, \mathcal{D}_i^{\text{test}})$. Mathematically, it can be stated as
\begin{equation*}
    \boldsymbol\theta^*_{\text{ML}}:= \underset{\boldsymbol\theta \in \boldsymbol\Theta}{\text{argmin }} F(\theta),
\end{equation*}
\begin{equation}
\text{where, }F(\boldsymbol\theta) = \frac{1}{M}\sum_{i=1}^M\mathcal{L}\left(\mathcal{A}d(\boldsymbol\theta, \mathcal{D}_i^{\text{tr}}), \mathcal{D}^{\text{test}}_i\right).
\label{eq_meta}
\end{equation}
During inference, dataset $D^{\text{tr}}_j$ corresponding to a new task $\mathcal{T}_j \sim P(\mathcal{T})$ is used to update $\boldsymbol\theta_{\text{ML}}$ to obtain the task-specific parameters $\boldsymbol\phi_j$. This is further denoted as \mbox{$\boldsymbol\phi_j = \mathcal{A}d(\boldsymbol\theta_{\text{ML}}, D^{\text{tr}}_j)$}.

\subsection{Kernel selection in meta-learning}
\begin{figure}
    \centering
\begin{subfigure}{0.5\textwidth}
\centering
    \includegraphics[width=0.85\textwidth]{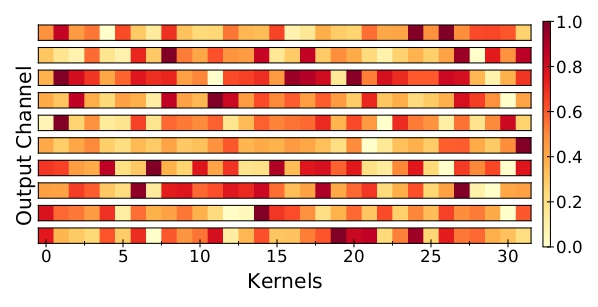}
    \vspace{-0.5em}
    \caption{Without budget constraint}
    \label{fig_a_kernel}
\end{subfigure}
\begin{subfigure}{0.5\textwidth}
\centering
    \includegraphics[width=0.85\textwidth]{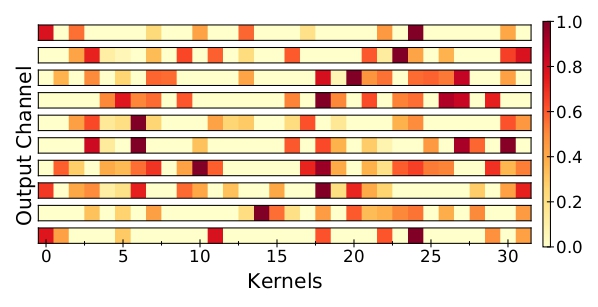}
    \vspace{-0.5em}
    \caption{With budget constraint}
    \label{fig_b_kernel}
\end{subfigure}
\vspace{-0.1in}
    \caption{Relative contribution of 32 different $3\times3$ kernels in mapping 10 different output channels for (a) no budget constraint, and (b) a budget constraint of 50\% on the total number of kernels that are used per mapping.}
    \label{fig_kernel}
    \vspace{-0.15in}
\end{figure}

\begin{figure}
    \centering
    \includegraphics[width=0.45\textwidth]{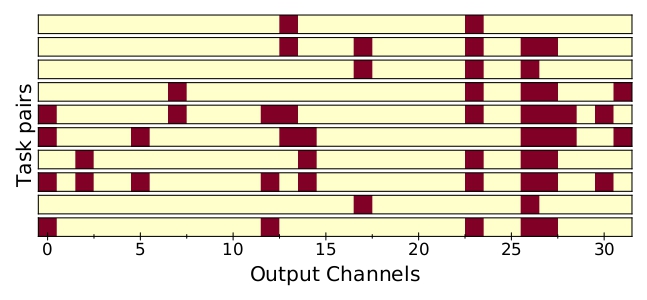}
    \vspace{-1em}
    \caption{Binary correlation map of construction of 32 different output channels for 10 randomly sampled task pairs. Here, `yellow' indicates that the respective output channel for both tasks in the corresponding task pair was constructed using exactly the same set of input kernels and `red' indicates otherwise.}
    \vspace{-0.1in}
    \label{fig-task-pruning}
    \vspace{-0.15in}
\end{figure}

As has been described above, the traditional approach in meta-learning is to learn a meta-model that can be adapted to different tasks. Although the tasks are assumed to be sampled from a common distribution $P(\mathcal{T})$, they vary to some extent. Thus, it is evident that for a single meta-model to fit across all the tasks, it needs to possess sufficient bandwidth of model weights. However, it is of interest to explore whether all the model weights are indeed needed for each of these tasks, and if not, can we adapt this usage in a task-specific manner. 

Figure \ref{fig_kernel} shows an example of kernel usage for a few mappings between the kernels and the output channels. For the sake of visual clarity, we show here only first 32 kernels of the total 64 from the last layer of our model. For each output channel, we show the relative contribution of the various kernels. From Figure \ref{fig_a_kernel}, it is seen that the contributions of the different input kernels vary across different output channels with some kernels being more important than the others for an output channel. Figure \ref{fig_b_kernel} shows the same 32 kernels as above, but for a model pruned with our method to a budget constraint of 50\% on the total fraction of the kernels to be used. The resulting sparsity is clearly visible in \mbox{Figure \ref{fig_b_kernel}} where we see that for each output channel, a significant set of kernels are no more relevant. We have observed and later report in this paper that even at such sparse configurations, the overall performance of the model is at par with the model trained without any budget constraint. As anticipated, different output channels use different set of input kernels, however, the overall selection of kernels is still sparse. Clearly, there is a scope of using a reduced number of kernels per output channel, and we achieve it in this paper through smart kernel selection, also referred as kernel pruning. 

Kernel selection, when combined with meta-learning, facilitates the identification of optimal set of kernels specific to each task. This leads to compressed meta-models as well as task-specific models that are less vulnerable to overfitting and provide improved memory efficiency. Figure \ref{fig-task-pruning} shows an example of kernel selection used for task-specific pruning in meta-learning. We see here the binary correlation map for multiple task pairs, and it is observed that the choice of input kernels for constructing the output channels varies across the tasks. Clearly, not all kernels of a model are relevant for each task, and the unused kernels can be eliminated through our dynamic kernel selection strategy.

\subsection{\textsc{MetaDOCK} formulation}
We present here \textsc{MetaDOCK}, a dynamic kernel selection strategy for meta-learning that treats task-specific kernel selection as an integral part of the model optimization process. To start with, \textsc{MetaDOCK} modifies the optimization problem stated in Eq. \ref{eq_meta} as follows.
\begin{equation*}
    \boldsymbol\theta^*_{\text{ML}}, \mathbf{z}^*_{\text{ML}}:= \underset{\boldsymbol\theta \in \boldsymbol\Theta, \mathbf{z} \in \mathcal{Z}}{\text{argmin }} F(\boldsymbol\theta, \mathbf{z}),
    \vspace{-0.5em}
\end{equation*}
\vspace{-1em}
\begin{equation}
    F(\boldsymbol\theta,\mathbf{z})=\frac{1}{M}\sum_{i=1}^M\mathcal{L}\left(\mathcal{A}d(\mathbf{z}, \boldsymbol\theta; \mathcal{D}_i^{\text{tr}}), \mathcal{D}^{\text{test}}_i\right).
\label{eq_meta2}
\end{equation}
Note here that $\mathbf{z}$ denotes a set of masking parameters. Based on this formulation, it is evident that the pruning masks are learnt at two different levels: meta-level and adaptation phase (inner-level). The inner step facilitates to identify kernels that are to be used for a certain task and the rest are ignored.

For the inner-level learning process, \textsc{MetaDOCK} employs the adaptation function $\mathcal{A}d(\cdot)$. Inspired from the iMAML algorithm, we construct $\mathcal{A}d(\cdot)$ to be the following regularized problem:
\begin{align}
    \mathcal{A}d^*(\boldsymbol\theta, \mathbf{z}; \mathcal{D}^{\text{tr}}_i) = & \underset{\boldsymbol\phi \in \boldsymbol\Theta, \boldsymbol\zeta \in \mathcal{Z}}{\text{argmin}} \mathcal{L}(\mathcal{B}(\boldsymbol\zeta)\odot\mathcal{K}(\boldsymbol\phi); \mathcal{D}^{\text{tr}}_i)  \nonumber \\ 
     & +  \frac{\lambda_1}{2}\|\boldsymbol\theta - \boldsymbol\phi\|^2 +  \frac{\lambda_2}{2}\|\mathbf{z} - \boldsymbol\zeta\|^2 \nonumber \\
     &+ \frac{\lambda_3}{2}\|\mathcal{V}(\boldsymbol\zeta)-\mathcal{V}_0\|^2 + \lambda_4\| \boldsymbol\zeta \|
     \label{eq_metadock1}
\end{align}

The task-specific model parameters and the pruning masks are denoted by $\boldsymbol\phi$ and $\boldsymbol\zeta$ in the equation above.
Further, $\mathcal{K}(\boldsymbol\phi)$ denotes the set of kernels constructed from the model parameters $\boldsymbol\phi$ and $\odot$ denotes the elementwise multiplication operation. Each mask gets multiplied with one kernel in the network such that the total count of masks is equal to the number of kernels of the model. When masking the kernels, the masks $\boldsymbol\zeta$ are projected to 0 or 1 using the operator $\mathcal{B}(\cdot)$, and this function is defined as 
\begin{equation}
  \mathcal{B}(\zeta_j) = \left\{
  \begin{array}{@{}ll@{}}
    1, & \text{if}\ \zeta_i > 0 \\
    0, & \text{otherwise}
  \end{array}\right.
  \label{binarize}
\end{equation}

As seen in Eq. \ref{eq_metadock1}, there are four regularization terms in the objective function weighted by terms $\lambda_1$, $\lambda_2$, $\lambda_3$ and $\lambda_4$. Similar to the conventional iMAML strategy, the first term encourages the task-specific model parameters $\boldsymbol\phi_i$ to stay close to $\boldsymbol\theta$. This regularization helps to obtain a strong prior on $\boldsymbol\theta$ such that the learnt meta-model requires only a very few update steps and very limited new training samples to adapt to a new task. Similarly, the second regularization term in Eq. \ref{eq_metadock1} ensures that the task-specific dynamic kernel selection builds up on the kernels identified to be active at the meta-model level. This regularization imposes a strong prior on $\mathbf{z}$ and helps to dynamically select the task-specific kernels with only a few update steps. The third regularization terms accounts for the budget constraint to be imposed on the optimization problem. The total budget $\mathcal{V}(\boldsymbol\zeta)$ for any task is defined as
\begin{equation}
    \mathcal{V}(\boldsymbol\zeta) = \frac{\text{Sum total of active kernels, }\sum \mathcal{B}(\zeta_j)}{\text{Total number of kernels}}.
\end{equation}
For $M$ tasks, we have $M$ such inequality constraints, and rather than incorporating them directly, we add a combined regularization term for the same. The implementation of budget constraint is generic in our implementation, and can be used to impose budget on model volume, channels and FLOPs, among others, as in \cite{tiwari2021chipnet, lemaire2019structured}. The last regularization term induces $\ell_{1}$-sparsity on $\mathbf{z}$. For the masks associated with most dynamic kernels, we have observed that this penalty term helps to keep the value of $z$ close to 0, thereby facilitating easy switching between active and inactive across different tasks. 

\subsection{\textsc{MetaDOCK}  optimization}
\begin{figure*}
    \centering
    \includegraphics[width=0.9\textwidth]{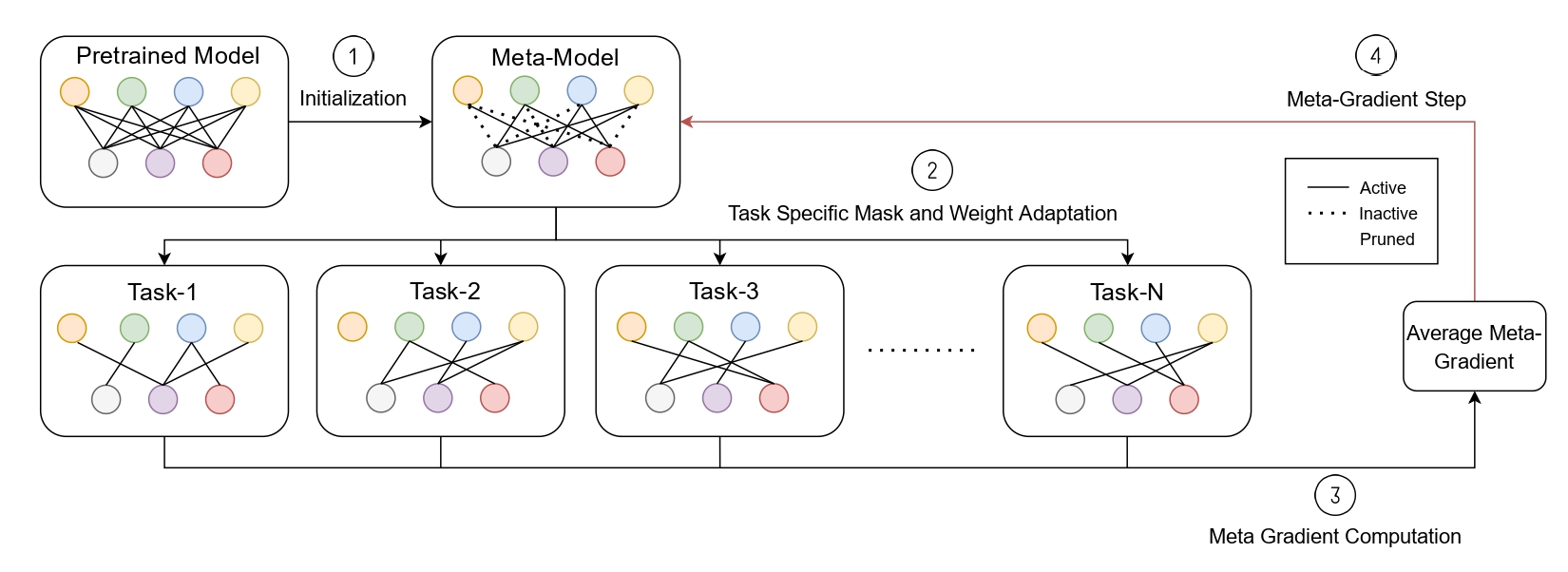}
    \vspace{-2em}
    \caption{Schematic representation of the \textsc{MetaDock} pipeline.}
    \label{fig:metadockopt}
    \vspace{-0.1in}
\end{figure*}

The optimization problem posed in Eq. \ref{eq_metadock1} is solved through iteratively alternating between the following two update steps:
\begin{align}
    \boldsymbol\theta \leftarrow \boldsymbol\theta -  \eta_1 d_{\boldsymbol\theta}F(\boldsymbol\theta, \mathbf{z}; \mathcal{D}_i^{\text{tr}}), \label{eq_update_theta}\\
    \mathbf{z} \leftarrow \mathbf{z} - \eta_2 d_{\mathbf{z}}F(\boldsymbol\theta, \mathbf{z}; \mathcal{D}_i^{\text{tr}}), \label{eq_update_z}
\end{align}
where $d_{\boldsymbol\theta}(\cdot)$ and $d_{\mathbf{z}}(\cdot)$ denote the first-order gradients with respect to $\boldsymbol\theta$ and $\mathbf{z}$, respectively. Based on Eq. \ref{eq_metadock1}, the update step for the masks can be further expanded as

\begin{equation}
      \mathbf{z} \leftarrow \mathbf{z} -  \eta_2 \frac{1}{M}\sum_{i=1}^M\frac{d\mathcal{A}d^*_i(\boldsymbol\theta, \mathbf{z})}{d\boldsymbol\zeta}\nabla_{\mathbf{z}}\mathcal{L}\left(\mathcal{A}d^*(\mathbf{z}, \boldsymbol\theta)\right). 
\end{equation}
A similar expression can be stated for $\boldsymbol\theta$. This two-step update strategy allows to achieve stability of the inter-twinned optimization problem as well as helps convergence with the gradient descent method. 

Figure \ref{fig:metadockopt} shows the schematic representation of the workflow of \textsc{MetaDOCK} and additional details on the pipeline of the approach are provided in Algorithm \ref{algo_metadock}. \textsc{MetaDOCK} starts with initializing the meta-model with weights from a pretrained model, thus $\boldsymbol\theta = \boldsymbol\theta_0$. With every kernel of the initial model, we then associate a pruning mask, and the values of the mask are sampled from a uniform distribution (to keep all the masks active initially). The goal of \textsc{MetaDOCK} is then to simultaneously optimize the model weights as well as the associated masks. At every meta-step, a set of tasks are sampled from the pre-defined distribution of tasks, and inner updates are performed independently for each task to obtain $\boldsymbol\phi_i$ from $\boldsymbol\theta$. Based on the average of the error gradient across the whole batch of tasks, an update to the meta-model is performed. This process is repeated until the desired level of convergence is achieved. Details related to each function outlined in Algorithm \ref{algo_metadock} can be found in the supplementary material.

For better convergence, $\textsc{MetaDOCK}$ employs a two step model update strategy as stated in Eqs. \ref{eq_update_theta} and \ref{eq_update_z}. In the first step, model is updated with the gradients of the error with respect to the model parameters while keeping the values of $\mathbf{z}$ thresholded to 0 or 1 using $\mathcal{B}(\cdot)$. Note that due to $\mathcal{B}(\cdot)$, gradients cannot be computed for $\mathbf{z}$. Thus, during the second step, model is updated using pseudo-gradients, \emph{i.e.}, the backward propagation ignores the projection function and the gradient is approximated with a sigmoid function. This approach is very commonly employed in the pruning and binarization literature (\emph{e.g.} \cite{DBLP:journals/corr/abs-1808-00278, Qin:cvpr20}), and we have found it to work well in the \textsc{MetaDOCK} framework as weights are directly optimized over discrete masks eliminating the need of separate retraining stage completely.

\begin{algorithm}[!t]
\small{
\DontPrintSemicolon

\KwIn{Distribution of tasks: $P(\mathcal{T})$, Pre-trained model parameters: $\boldsymbol\theta_{0}$, Budget: $V_{0}$, Meta iterations: $N_{\text{outer}}$, Meta batch size: $B$, Inner iterations: $N_{\text{inner}}$, Scalars: $\lambda_{1}$, $\lambda_{2}$, $\lambda_{3}$, $\lambda_{4}$}
\KwOut{Parameters: $\boldsymbol \theta^*$, Masks: $\mathbf{z}^*$} 
\Initialize{Model weights: $\boldsymbol  \theta=\boldsymbol\theta_{0}$\\
kernel masks: $\mathbf{z} \in \mathcal{U}(0, 0.01)$ \;}
\SetKwBlock{Begin}{function}{end function}
{
  {
    \For{$i = 1, 2, ... N_{\text{outer}} $}
    {
        Sample mini-batch of tasks $\left\{\mathcal{T}_{i}\right\}_{i=1}^{B} \sim P(\mathcal{T})$ \;
        \For{each task $\mathcal{T}_{i}$}
        {
            $\boldsymbol \phi = \boldsymbol \theta, \boldsymbol \zeta = \mathbf{z}$ \;
            \For{$j = 1, 2, ... N_{\text{inner}}$}
            {
                $\boldsymbol\zeta_{b} \gets \mathcal{B}(\boldsymbol\zeta)$ \;
                $\hat{\mathbf{y}} \gets \texttt{Forward}(\mathbf{x}, \boldsymbol \phi, \boldsymbol\zeta_{b})$ \;
                $\mathcal{L}_{1} \gets \texttt{MetaLoss}(\mathbf{y}, \hat{\mathbf{y_{1}}}, \lambda_{1}, \lambda_{2}, \boldsymbol \phi, \boldsymbol \theta, \boldsymbol\zeta, \mathbf{z})$ \;
                $\mathcal{L}_{2} \gets  \mathcal{L}_{1}+ \lambda_{3}\texttt{Bud.Loss}(\boldsymbol\zeta_{b}, V_{0}) + \lambda_{4}\lVert \boldsymbol \zeta \rVert $\;
                $\nabla \boldsymbol \phi \gets \texttt{Backward}(\mathcal{L}_{1})$ \;
                $\nabla \boldsymbol\zeta \gets \texttt{PseudoBackward}(\mathcal{L}_{2}, \boldsymbol\zeta)$ \;
                $(\phi, \boldsymbol\zeta) \gets \texttt{OptimizeStep}(\nabla \boldsymbol \phi, \nabla \boldsymbol\zeta, \beta)$
            }
            Compute task meta gradient of $\boldsymbol \phi : g_{i}$, $\boldsymbol\zeta : h_{i}$ \;
        }
        $\hat{\nabla} G(\boldsymbol \theta)=(1 / B) \sum_{i=1}^{B} g_{i}$ \;
        $\hat{\nabla} H(\mathbf{z})=(1 / B) \sum_{i=1}^{B} h_{i}$ \;
        $\boldsymbol \theta \leftarrow \boldsymbol \theta-\eta \hat{\nabla} G(\boldsymbol \theta)$, 
        $\mathbf{z} \leftarrow \mathbf{z}-\eta \hat{\nabla} H(\mathbf{z})$
    }
  }
}
\caption{\textsc{MetaDOCK} for iMAML}
\label{algo_metadock}
\vspace{-0.1in}
}
\end{algorithm}
\section{Experiment}

We discuss here various experiments conducted in this paper to demonstrate the efficacy of \textsc{MetaDOCK}. First we conduct an experiment to analyze how our two-step pruning strategy performs compared to the classical continuous pruning for meta-learning. In another set of experiments, we perform a comparative study between task-specific pruning as well as global pruning. It is of interest to analyze whether the few inner update steps are sufficient to perform task-specific pruning, and if yes, we study whether the resultant models perform well or not. We propose a metric to measure the extent of overfit in meta-learning and conduct experiments to analyze whether the models obtained from \textsc{MetaDOCK} generalize better than their unpruned counterparts. 

%We then show that task specific pruning extract better distinct target models for each task as compared to general global pruning which extracts a single meta-model for all tasks. 
We study the performance of \textsc{MetaDOCK} on standard 4-conv models trained on mini-ImageNet \cite{vinyals2016matching} and CIFAR-fs \cite{bertinetto2018meta} datasets for several different choices of pruning budget. For the meta-validation tasks,  we sample 600 tasks randomly from the meta-validation split.
%We then present task-specific experiments on standard 4-conv models, mini-ImageNet \cite{vinyals2016matching} and CIFAR-fs \cite{bertinetto2018meta} datasets at different target budgets of pruning. 
%We also show that pruning improves generalization of meta-learning methods with the help of our designed Measure of Overfit (O). We sample 600 tasks randomly from the meta-validation split to form our meta-validation tasks. 
In contrast to the existing methods, we take all possible combinations of classes from meta-test split to generate a large number of meta-test tasks. For $C$ classes in the complete meta-test set, the total number of meta-test tasks we generate for N-way setting is $\Comb{C}{N}$. This translates to 15504 test tasks for both CIFAR-fs and mini-ImageNet $(C=20, N=5)$. 

We evaluate on meta-validation tasks after every 1000 meta-training steps and save the model based on the best average meta-validation accuracy. Finally, we evaluate this saved model on all generated meta-test tasks. This strategy is much more robust than reporting meta-validation accuracy alone which often overfits due to repeated evaluation during training. We set $\lambda_{3}$ (budget weight) to 50 and $\lambda_{4}$ ($l_{1}$ weight) to $10^{-6}$ in all our pruning experiments. Note that for the extreme pruning scenarios, that involve very low budgets, larger value of $\lambda_{3}$ should be chosen to satisfy the target budget, however, we prioritize performance over satisfying the budget strictly and thus chose a fixed value of $\lambda_{3}$ across all budgets. Lastly, $\lambda_{1}$ is set to 0.5 as reported originally in \cite{rajeswaran2019meta}.  Since $\lambda_{2}$ has a similar regularization effect as $\lambda_{1}$ but on masks, hence we set it to 0.5 as well. We show experimental results at target budgets of 50\%, 25\%, 12.5\% and 6.25\%. Lastly, we use the same optimization hyperparameters for pruning which were used during the pre-training stage in \cite{rajeswaran2019meta}.

\subsection{Optimal pruning strategy for meta-learning}
\begin{table}[]
\centering
\resizebox{0.48\textwidth}{!}{
\begin{tabular}{cc|cc}
\textbf{Budget}        & \textbf{Method} & \textbf{Val Acc (\%)} & \textbf{Test Acc (\%)}   \\ \hline
25.0 & Continuous          &  67.49 ± 1.00          & 67.78 ± 0.19                         \\
25.2   & \textsc{MetaDOCK}   &  69.65 ± 0.94        &  69.30 ± 0.19                          \\ \hline
12.5 & Continuous          &  64.40 ± 1.01            &  64.87 ± 0.20           \\
13.0 & \textsc{MetaDOCK}  &  70.69 ± 0.97        &  70.15 ± 0.19                       
\end{tabular}
}
\vspace{-0.1in}
\caption{Performance scores  on CIFAR-fs dataset, 5-way 5-shot setting for continuous pruning and discrete two-step (\textsc{MetaDOCK}) pruning at different budgets.}
\label{tab:contdisc}
\vspace{-0.15in}
\end{table}

\noindent{\textbf{Continuous vs. two-step model update scheme}}. Commonly, most pruning methods employ continuous pruning scheme. This involves learning soft masks as part of the model training process. Rather than using discrete values of 0 and 1 for the masks, continuous schemes optimize them for [0, 1], and add additional penalty on the objective function to push final masks to 0 (inactive kernel) and 1 (active kernel). As described earlier, \textsc{MetaDOCK} employs a two-step discrete pruning scheme and in this experiment, we compare its performance with the classical continuous pruning method. 

For the continuous scheme, we follow the approach similar to \cite{Liu2017iccv}, and introduce kernel masks to induce sparsity in the pre-trained meta model. The target objective is to rank the kernels according to their influence on the final performance with the help of induced mask values. This model is jointly trained with regular meta-learning loss functions combined with $\ell_{1}$-norm on masks with the same training hyperparameters as used in pre-training stage. Once pruning is completed, depending upon target budget, kernels with low mask values are eliminated and compressed meta-model structure is extracted. Finally, this compressed meta-model is retrained with similar strategy used for pre-training so that the weights can adjust themselves from continuous masks in the pruning stage to binary/non-existing masks in the final structure. An important point to note is that models pruned with this strategy has a reduced size but it is task-independent, i.e., the same meta-model is used in all tasks. In contrast to our two step discrete pruning strategy where weight optimization on binary masks occur in a single stage, continuous pruning requires an additional finetuning stage for the compressed meta-model to adapt model weights from continuous masks in pruning stage to discrete/non-existing masks in final model thus increasing total pruning time.

The results for continuous scheme as well as our approach on CIFAR-fs, 5-way, 5-shot are shown in Table \ref{tab:contdisc}. It can be clearly seen that our pruning method outperforms the continuous pruning baseline at both 25\% and 12.5\% budgets. At extreme budget (12.5\%), our approach improves the performance significantly over the baseline. This clearly indicates that the two-step strategy with discrete projection employed in \textsc{MetaDOCK} is more stable and a better approach for pruning in meta-learning.

\begin{table}[]
\centering
\resizebox{0.48\textwidth}{!}{
\begin{tabular}{cc|cc}
\textbf{Budget}        & \textbf{Method} & \textbf{Val Acc (\%)} & \textbf{Test Acc (\%)}   \\ \hline
13.9   & Global          &  69.95 ± 0.98        &  69.98 ± 0.19            \\
13.0   & Task-Specific   &  70.69 ± 0.97        &  70.15 ± 0.19                          \\ \hline
7.8    & Global          &  69.96 ± 0.98        &  69.92 ± 0.19           \\
7.2    & Task-Specific   &  70.49 ± 0.99        &  70.21 ± 0.19                      
\end{tabular}
}
\vspace{-0.1in}
\caption{Performance of CIFAR-fs, 5-way 5-shot setting for global and task-specific pruning at different budgets}
\label{tab:globaltask}
\vspace{-0.15in}
\end{table}

\noindent{\textbf{Global pruning vs. task-specific pruning}.} We further analyze if our task-specific pruning adds value to the overall performance and efficiency of the model. For this, we compare our approach with global pruning, where only pruning is performed at the meta-model level. Task specific pruning allows each task to adapt the model weights as well as mask values, thereby changing the architectures accordingly. However, global pruning freezes the compressed meta-model and only allows weights to adapt task-wise. 
%We further enforce budget loss and $\ell_{1}$-Norm on masks along with meta-learning loss in the meta-testing stage of task-specific pruning, hence task-specific pruning achieves a greater degree of compression as compared to global pruning where only meta-learning loss is employed.

The results for global and task-specific pruning on CIFAR-FS, 5-way 5-shot are shown in Table \ref{tab:globaltask}. Task-specific pruning outperforms global pruning at both 12.5\% and 6.25\% budgets while achieving a greater degree of compression at the same time. This asserts the fact that based on the difficulty of the task, model should be able to compress or expand itself in a given range. This further motivates us to adopt task-specific pruning in all our further experiments. 

\subsection{Learning task-specific compressed meta-models}

\begin{table*}
\centering
\small
\begin{minipage}{0.45\textwidth}
\centering
\begin{tabular}{cc|cc|c}
\multicolumn{2}{c|}{\textbf{Budget (\%)}} & \multicolumn{2}{c|}{\textbf{Accuracy (\%) $\uparrow$}} & \multirow{2}{*}{\textbf{\begin{tabular}[c]{@{}r@{}}Measure of \\ Overfit (MO)$\downarrow$\end{tabular}}} \\
\textbf{Meta}       & \textbf{Task}       & \textbf{Validation}     & \textbf{Test}     &                                                  \\ \midrule
\multicolumn{2}{c|}{iMAML~\cite{rajeswaran2019meta}\footnotemark}                                   & 68.97 ± 0.94            & 68.08 ± 0.19      &  15.5                                                \\
54.3                & 50.2                & 69.45 ± 0.96            & 68.62 ± 0.19      &  18.2                                                \\
28.0                & 25.2                & 69.65 ± 0.94            & 69.30 ± 0.19      &  14.1                                                \\
13.9                & 13.0                & 70.69 ± 0.97            & 70.15 ± 0.19      &  9.7                                                \\
7.8                 & 7.2                 & 70.49 ± 0.99            & 70.21 ± 0.19      &  6.1                                                \\
4.8                 & 4.1                 & 68.85 ± 0.97            & 67.88 ± 0.19      &  3.9                     

\end{tabular}
\vspace{-0.1in}
\caption{Performance of 4-Conv model with 128 channels on CIFAR-fs with 5-way 5-Shot setting at different budgets. \\}
\label{tab:4conv2cifar5w5s}
% \vspace{-0.15in}
\end{minipage}
\hspace{2em}
\begin{minipage}{0.45\textwidth}
\centering
\begin{tabular}{cc|cc|c}
\multicolumn{2}{c|}{\textbf{Budget (\%)}} & \multicolumn{2}{c|}{\textbf{Accuracy (\%) $\uparrow$}} & \multirow{2}{*}{\textbf{\begin{tabular}[c]{@{}r@{}}MO $\downarrow$ \end{tabular}}} \\
\textbf{Meta}       & \textbf{Task}       & \textbf{Validation}     & \textbf{Test}     &                                                  \\ \midrule
\multicolumn{2}{c|}{iMAML~\cite{rajeswaran2019meta}}                 & 56.50 ± 1.98            & 55.23 ± 0.38      &   35.4                                               \\
53.0                & 50.4                & 58.27 ± 1.91            & 55.46 ± 0.38      &   30.7                                               \\
26.3                & 25.7                & 58.40 ± 1.89            & 56.04 ± 0.38      &   24.4                                               \\
13.9                & 13.7                & 57.27 ± 1.85            & 55.38 ± 0.38      &   20.5                                               \\
7.4                 & 7.3                 & 57.20 ± 1.91            & 55.72 ± 0.38      &   15.5                                               \\
4.2                 & 4.1                 & 56.82 ± 1.95            & 53.27 ± 0.38      &   13.1

\end{tabular}
\vspace{-0.1in}
\caption{Performance of 4-Conv model with 128 channels on CIFAR-fs with 5-way 1-Shot setting at different budgets. \\}
\label{tab:4conv2cifar5w1s}
% \vspace{-0.15in}
\end{minipage}
\vspace{2em}
\begin{minipage}{0.45\textwidth}
\centering
\begin{tabular}{cc|cc|c}
\multicolumn{2}{c|}{\textbf{Budget (\%)}} & \multicolumn{2}{c|}{\textbf{Accuracy (\%) $\uparrow$}} & \multirow{2}{*}{\textbf{\begin{tabular}[c]{@{}r@{}}MO $\downarrow$ \end{tabular}}} \\
\textbf{Meta}       & \textbf{Task}       & \textbf{Validation}     & \textbf{Test}     &                                                  \\ \midrule
\multicolumn{2}{c|}{iMAML~\cite{rajeswaran2019meta}}                 & 67.90 ± 0.98            & 67.25 ± 0.20      &  6.0                                                \\
53.0                & 50.2                & 69.51 ± 0.94            & 69.17 ± 0.19      &  8.6                                                \\
26.9                & 25.2                & 69.91 ± 0.96            & 69.75 ± 0.19      &  4.8                                               \\
14.1                & 12.8                & 68.35 ± 0.97            & 67.92 ± 0.19      &  3.6                                                \\
7.2                 & 6.8                 & 66.66 ± 0.98            & 66.53 ± 0.19     &  2.9                  

\end{tabular}
\vspace{-0.1in}
\caption{Performance of 4-Conv model with 64 channels on CIFAR-fs with 5-way 5-Shot setting at different budgets.}
\label{tab:4conv1cifar5w5s}
\vspace{-0.15in}

\end{minipage}
\hspace{2em}
\begin{minipage}{0.45\textwidth}
\centering
\begin{tabular}{cc|cc|c}
\multicolumn{2}{c|}{\textbf{Budget (\%)}} & \multicolumn{2}{c|}{\textbf{Accuracy (\%) $\uparrow$}} & \multirow{2}{*}{\textbf{\begin{tabular}[c]{@{}r@{}}MO $\downarrow$ \end{tabular}}} \\
\textbf{Meta}       & \textbf{Task}       & \textbf{Validation}     & \textbf{Test}     &                                                  \\ \midrule
\multicolumn{2}{c|}{iMAML~\cite{rajeswaran2019meta}}                 & 55.97 ± 1.95            & 54.10 ± 0.38      &  24.7                                                \\
50.7                & 50.0                & 56.30 ± 2.02            & 54.55 ± 0.38      &  22.3                                                \\
25.7                & 25.2                & 57.33 ± 1.91            & 55.28 ± 0.38      &  14.7                                                \\
13.1                & 12.8                & 56.40 ± 1.97            & 54.02 ± 0.38      &  11.3                                                \\
7.7                 & 7.6                 & 55.37 ± 1.92            & 52.37 ± 0.38      &  9.5

\end{tabular}
\vspace{-0.1in}
\caption{Performance of 4-Conv model with 64 channels on CIFAR-fs with 5-way 1-Shot setting at different budgets.}
\label{tab:4conv1cifar5w1s}
\vspace{-0.15in}
\end{minipage}
\vspace{-0.2in}
\end{table*}
\footnotetext{iMAML~\cite{rajeswaran2019meta} numbers as baseline reproduced using the official implementation \url{https://github.com/aravindr93/imaml_dev}}

We discuss here the results of task-specific pruning obtained using 
\textsc{MetaDOCK} on standard 4-Conv models with 64 and 128 channels for CIFAR-fs 5-way 1-shot and 5-way 5-shot settings. Related results are presented in Tables \ref{tab:4conv2cifar5w5s}, \ref{tab:4conv2cifar5w1s}, \ref{tab:4conv1cifar5w5s} and \ref{tab:4conv1cifar5w1s}. Here, meta-budget indicates the fraction of kernels retained in the meta-model, and task-budget refers to the fraction remaining in the task-specific models. In general, we see that the performance of the pruned meta-models is higher than their unpruned counterparts and this gain is observed to be the highest for 5-way, 5-shot setting with 128 channels (Table \ref{tab:4conv2cifar5w5s}). For this case, pruning greatly improves over the validation and test sets and the best performance is achieved at $\sim$7\% budget increasing the performance over the base model by $\sim$2\%. Similarly in 5-way, 1-shot setting with 128 channels (Table \ref{tab:4conv2cifar5w1s}) pruning at 25\% budget improves the base performance by $\sim$1\%. 

\begin{table*}
\centering
\small
\begin{minipage}{0.45\textwidth}
\centering
\begin{tabular}{cc|cc|c}
\multicolumn{2}{c|}{\textbf{Budget (\%)}} & \multicolumn{2}{c|}{\textbf{Accuracy (\%) $\uparrow$}} & \multirow{2}{*}{\textbf{\begin{tabular}[c]{@{}r@{}}MO $\downarrow$ \end{tabular}}} \\
\textbf{Meta}       & \textbf{Task}       & \textbf{Validation}     & \textbf{Test}     &                                                  \\ \midrule
\multicolumn{2}{c|}{iMAML~\cite{rajeswaran2019meta}}                 & 61.55 ± 0.91            & 63.39 ± 0.18      &  17.6                                                \\
57.4                & 50.3                & 63.19 ± 0.91            & 63.54 ± 0.18      &  18.8                                                \\
32.2                & 25.3                & 63.65 ± 0.92            & 64.34 ± 0.18      &  13.1                                                \\
15.9                & 12.8                & 63.62 ± 0.94            & 64.05 ± 0.18      &  11.4                                                \\
8.2                 & 6.8                 & 63.32 ± 0.91            & 63.81 ± 0.18      &  8.9

\end{tabular}
\vspace{-0.1in}
\caption{Performance of 4-Conv model with 128 channels on mini-ImageNet with 5-way 5-Shot setting at different budgets.\\}
\label{tab:4conv2mini5w5s}
% \vspace{-0.15in}

\end{minipage}
\hspace{2em}
\begin{minipage}{0.45\textwidth}
\centering
\begin{tabular}{cc|cc|c}
\multicolumn{2}{c|}{\textbf{Budget (\%)}} & \multicolumn{2}{c|}{\textbf{Accuracy (\%) $\uparrow$}} & \multirow{2}{*}{\textbf{\begin{tabular}[c]{@{}r@{}}MO $\downarrow$ \end{tabular}}} \\
\textbf{Meta}       & \textbf{Task}       & \textbf{Validation}     & \textbf{Test}     &                                                  \\ \midrule
\multicolumn{2}{c|}{iMAML~\cite{rajeswaran2019meta}}                 & 46.90 ± 1.77            & 45.55 ± 0.36      &  21.3                                                \\
57.8                & 50.5                & 47.07 ± 1.84            & 45.72 ± 0.36      &  23.4                                                \\
28.6                & 26.0                & 47.17 ± 1.89            & 45.90 ± 0.36      &  20.2                                                \\
12.9                & 12.8                & 47.10 ± 1.93            & 45.97 ± 0.36      &  17.3                                                \\ 
7.1                 & 7.0                 & 46.81 ± 1.90            & 45.55 ± 0.36      &  17.0

\end{tabular}
\vspace{-0.1in}
\caption{Performance of 4-Conv model with 128 channels on mini-ImageNet with 5-way 1-Shot setting at different budgets. \\}
\label{tab:4conv2mini5w1s}
% \vspace{-0.15in}
\end{minipage}
\vspace{2em}
\begin{minipage}{0.45\textwidth}
\centering
\begin{tabular}{cc|cc|c}
\multicolumn{2}{c|}{\textbf{Budget (\%)}} & \multicolumn{2}{c|}{\textbf{Accuracy (\%) $\uparrow$}} & \multirow{2}{*}{\textbf{\begin{tabular}[c]{@{}r@{}}MO $\downarrow$ \end{tabular}}} \\
\textbf{Meta}       & \textbf{Task}       & \textbf{Validation}     & \textbf{Test}     &                                                  \\ \midrule
\multicolumn{2}{c|}{iMAML~\cite{rajeswaran2019meta}}                 & 60.58 ± 0.94            & 62.13 ± 0.18      &   12.4                                               \\
56.2                & 50.2                & 63.30 ± 0.93            & 63.5 ± 0.18      &  13.2                                                \\
30.0                & 25.2                & 63.29 ± 0.93            & 63.4 ± 0.19      &  11.2                                                \\
15.1                & 12.8                & 62.90 ± 0.96            & 62.72 ± 0.18      &  9.6                                                \\
8.2                 &  6.4                & 61.34 ± 0.92            & 61.23 ± 0.18    &   7.4                 

\end{tabular}
\vspace{-0.1in}
\caption{Performance of 4-Conv model with 64 channels on mini-ImageNet with 5-way 5-Shot setting at different budgets.}
\label{tab:4conv1mini5w5s}
\vspace{-0.15in}

\end{minipage}
\hspace{2em}
\begin{minipage}{0.45\textwidth}
\centering
\begin{tabular}{cc|cc|c}
\multicolumn{2}{c|}{\textbf{Budget (\%)}} & \multicolumn{2}{c|}{\textbf{Accuracy (\%) $\uparrow$}} & \multirow{2}{*}{\textbf{\begin{tabular}[c]{@{}r@{}}MO $\downarrow$ \end{tabular}}} \\
\textbf{Meta}       & \textbf{Task}       & \textbf{Validation}     & \textbf{Test}     &                                                  \\ \midrule
\multicolumn{2}{c|}{iMAML~\cite{rajeswaran2019meta}}                 & 45.20 ± 1.88            & 44.58 ± 0.36      &   23.1                                               \\
60.2                &   50.4              & 46.20 ± 1.92            & 44.88 ± 0.36      &   21.3                                               \\
26.6                & 25.5                & 45.72 ± 1.93            & 44.92 ± 0.36      &   20.5                                               \\
13.0                & 12.7                & 45.27 ± 1.91            & 44.87 ± 0.36      &   14.8                                               \\
6.8                 & 6.6                 & 44.83 ± 1.82            & 44.67 ± 0.36      &   12.2

\end{tabular}
\vspace{-0.1in}
\caption{Performance of 4-Conv model with 64 channels on mini-ImageNet with 5-way 1-Shot setting at different budgets.}
\label{tab:4conv1mini5w1s}
\vspace{-0.15in}
\end{minipage}
\vspace{-0.2in}
\end{table*}

To analyze the generaliziblity of \textsc{MetaDOCK} across other datasets, we also report results for mini-ImageNet in Tables \ref{tab:4conv2mini5w5s}, \ref{tab:4conv2mini5w1s}, \ref{tab:4conv1mini5w5s} and \ref{tab:4conv1mini5w1s}. We observe that for this dataset as well, the pruned models consistently outperform the baseline model, except for very low budgets of less than 10\%, where minor drops in performance are observed. Similar to CIFAR-fs, performance gains of up to 2\% in accuracy are observed for this dataset as well. Clearly, with \textsc{MetaDOCK}, it is possible to learn compressed models that perform better than the baseline models in meta-learning. 

An interesting observation across both the datasets is that performing task-specific pruning of a larger model is more beneficial than just using an unpruned model of the same size. Intuitively, this seems right since the pruned model will have a more optimized distribution of kernels compared to the unpruned one. For example, we observe for both the datasets, that the 4-conv model when trained with 128 channels and pruned for 25\% budget, attains higher accuracy than the pretrained 4-conv model comprising 64 channels - both these models have the same size. The former attains an average meta-test accuracy score of 70.15\% on CIFAR-fs 5-way 1-shot (Table \ref{tab:4conv2cifar5w5s}), while the latter scores 67.25\%, thus increasing the performance by an absolute margin of 2.9\%.  Similarly, the corresponding increase in accuracy for mini-ImageNet dataset is $\sim$2.2\% (Table \ref{tab:4conv2mini5w5s} and \ref{tab:4conv1mini5w5s}). Similar trends can also be observed for other model pairs as well across datasets and different model settings. This suggests that pruning a large network to get task-specific models is better than using a pretrained model of similar size.

\begin{figure}
    \centering
    \includegraphics[width=0.4\textwidth]{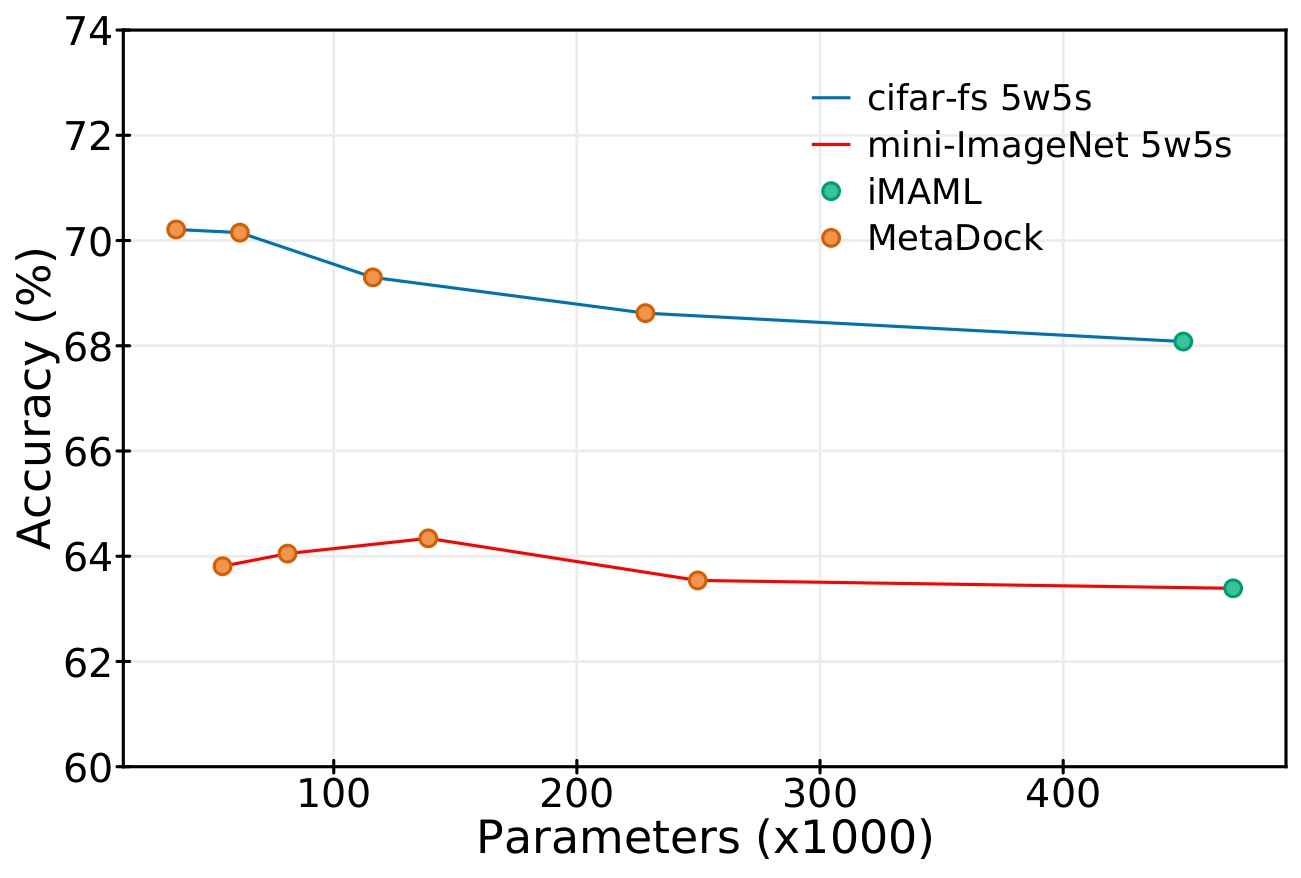}
    \vspace{-0.1in}
    \caption{Accuracy vs Parameters for 4-Conv 128 channels pruned with \textsc{MetaDock}.}
    \label{param}
    \vspace{-0.15in}
\end{figure}

\begin{figure}
    \centering
    \includegraphics[width=0.4\textwidth]{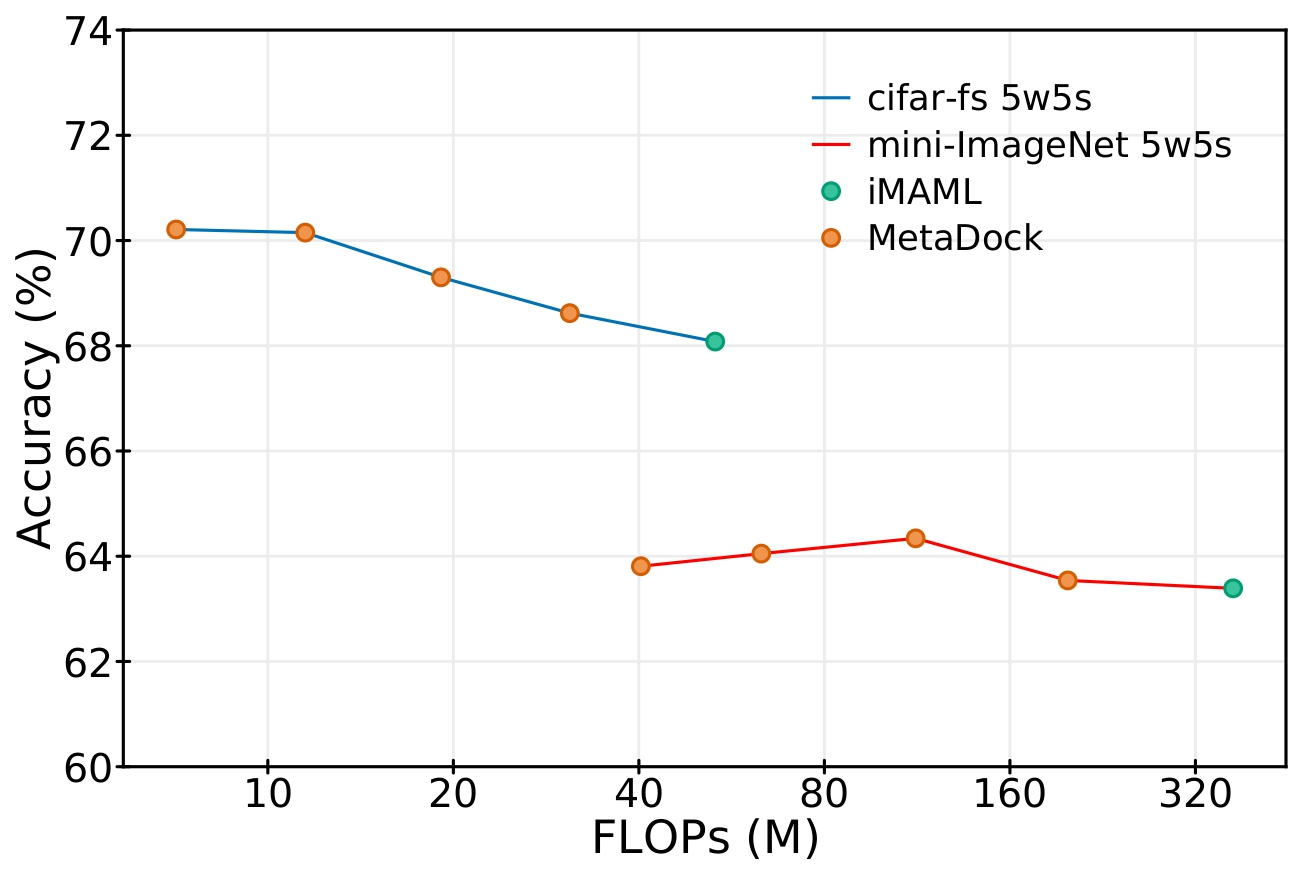}
    \vspace{-0.1in}
    \caption{Accuracy vs FLOPs for 4-Conv 128 channels pruned with \textsc{MetaDock}.}
    \label{flops}
    \vspace{-0.15in}
\end{figure}

We further study the compute and storage efficiency of our pruned models by analyzing the obtained accuracy scores versus the number of parameters and FLOPs at different pruning budgets. Figure \ref{param} shows parameters vs. accuracy plot for CIFAR-fs and mini-ImageNet datasets on 5-way, 5-shot setting with 4-Conv-128 model. We compare \textsc{MetaDOCK} with iMAML and it can be seen that \textsc{MetaDock} consistently outperforms iMAML while substantially reducing the number of parameters. After a threshold parameter count, performance starts to drop for both CIFAR-fs and mini-ImageNet but the extent of compression achieved before that point is already significant.

Similarly, Figure \ref{flops} shows FLOPs vs. accuracy plot for CIFAR-fs and mini-ImageNet dataset on 5-way, 5-shot setting with 4-Conv-128 model. For mini-ImageNet, more than $3\times$ reduction in FLOPs is achieved while improving the accuracy by 1\%. Similarly, for CIFAR-fs, more than $5\times$ reduction in FLOPs is achieved while improving the accuracy by more than 2\%. As seen in parameters vs. accuracy plot, performance here as well tends to drop beyond a certain FLOPs' threshold.

\noindent{\textbf{Pruning improves generalization}}. To measure the extent of generalization in meta-learning, we introduce a metric, \textit{Measure of Meta-Overfitting} (MO), to compare generalization of different models on unseen tasks. We define it as:
\begin{equation}
    \text{MO} = \frac{Acc_{\text{train}}- Acc_{\text{test}}}{Acc_{\text{test}}}\times 100
\end{equation}
where $Acc_{\text{train}}$ refers to the average accuracy on the test set of 600 randomly sampled tasks that are seen by the model during meta-training stage and $Acc_{\text{test}}$ refers to the average accuracy on the test set of 600 randomly sampled unseen tasks. A low value of this metric indicates that $Acc_{\text{train}}$ is close to $Acc_{\text{test}}$ and/or $Acc_{\text{test}}$ is fairly high which implies better generalization. Note that $\text{MO}$ alone cannot be used to judge a model's performance but it gives a fair measure of meta-overfitting. 

To analyze the measure of overfitting for \textsc{MetaDOCK}, we look at the \text{MO} scores reported in the tables above. It is observed that the task-specific pruned models show lower \textsc{MO} score and this score reduces with lower budget constraints. These results clearly demostrate that \textsc{MetaDOCK} helps to improve generalization in meta-learning.

\section{Conclusion and Future Work}
In this paper, we have presented $\textsc{MetaDOCK}$, a dynamic kernel pruning strategy to improve generalization and build memory efficient models in metalearning. $\textsc{MetaDOCK}$ achieves model compression at two levels - global/meta as well as task-specific. With its efficient adaptation capability, $\textsc{MetaDOCK}$ can create compressed models that generalize better on unseen tasks. Through several numerical experiments, we have demonstrated the efficacy of $\textsc{MetaDOCK}$. For our benchmark experiments, we demonstrated that $\textsc{MetaDOCK}$ builds pruned models whose accuracy scores are up to 2\% higher than the unpruned networks, while reducing the model size by more than 75\%. Further, we showed that for the same inference budget, pruned variants of larger models obtained from $\textsc{MetaDOCK}$ outperform the unpruned smaller networks substantially. With these results, we hope to have addressed an important research question of how metalearning models can generalize better on unseen tasks while achieving better memory efficiency. As part of our future work, we hope to integrate our method with other meta-learning strategies not limited to optimization-based method.

%Our method achieves two level of compressions - global and task specific. Models pruned with $\textsc{MetaDOCK}$ are compute- as well as space-efficient and can quickly adapt to a new task to further improve space and compute efficiency. These models have better performance and generalization for unseen tasks as compared to their pretrained counterparts. 

\clearpage

{\small
\bibliographystyle{ieee_fullname}
\bibliography{egbib}
}
\clearpage
% \include{appendix}
% Appendix

\end{document}